\documentclass[10pt, a4paper]{article}

\usepackage[final]{lrec-coling2024} 

\usepackage{graphicx}
\usepackage{adjustbox}
\usepackage{multirow}
\usepackage{stfloats}

\title{\textit{Historia Magistra Vitae}: Dynamic Topic Modeling of Roman Literature using Neural Embeddings}

\name{Michael Ginn and Mans Hulden} 

\address{University of Colorado \\
         \{michael.ginn, mans.hulden\}@colorado.edu\\}

\abstract{
Dynamic topic models have been proposed as a tool for historical analysis, but traditional approaches have had limited usefulness, being difficult to configure, interpret, and evaluate. In this work, we experiment with a recent approach for dynamic topic modeling using BERT embeddings. We compare topic models built using traditional statistical models (LDA and NMF) and the BERT-based model, modeling topics over the entire surviving corpus of Roman literature. We find that while quantitative metrics prefer statistical models, qualitative evaluation finds better insights from the neural model. Furthermore, the neural topic model is less sensitive to hyperparameter configuration and thus may make dynamic topic modeling more viable for historical researchers.
 \\ \newline  }

\begin{document}

\maketitleabstract

\section{Introduction}

\begin{figure*}[!b]
  \centering
  \begin{minipage}[b]{0.31\linewidth}
    \includegraphics[width=\linewidth]{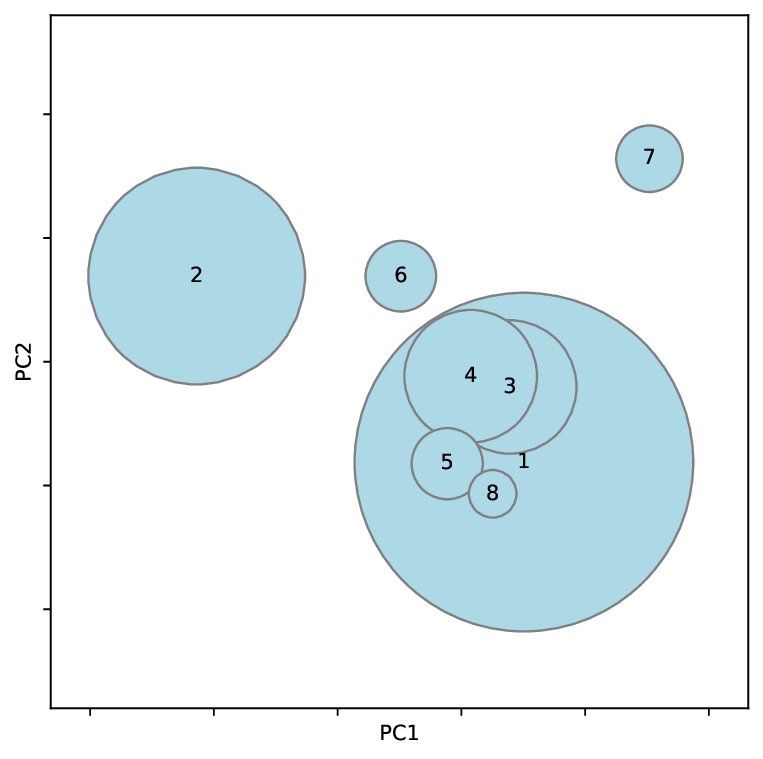}
  \end{minipage}
  \hfill
  \begin{minipage}[b]{0.31\linewidth}
    \includegraphics[width=\linewidth]{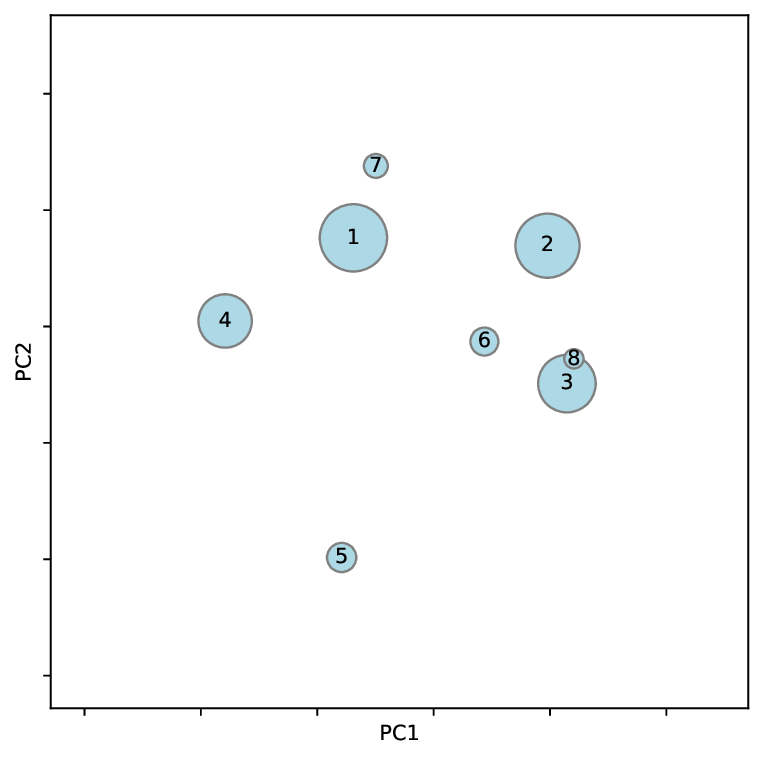}
  \end{minipage}
  \hfill
  \begin{minipage}[b]{0.31\linewidth}
    \includegraphics[width=\linewidth]{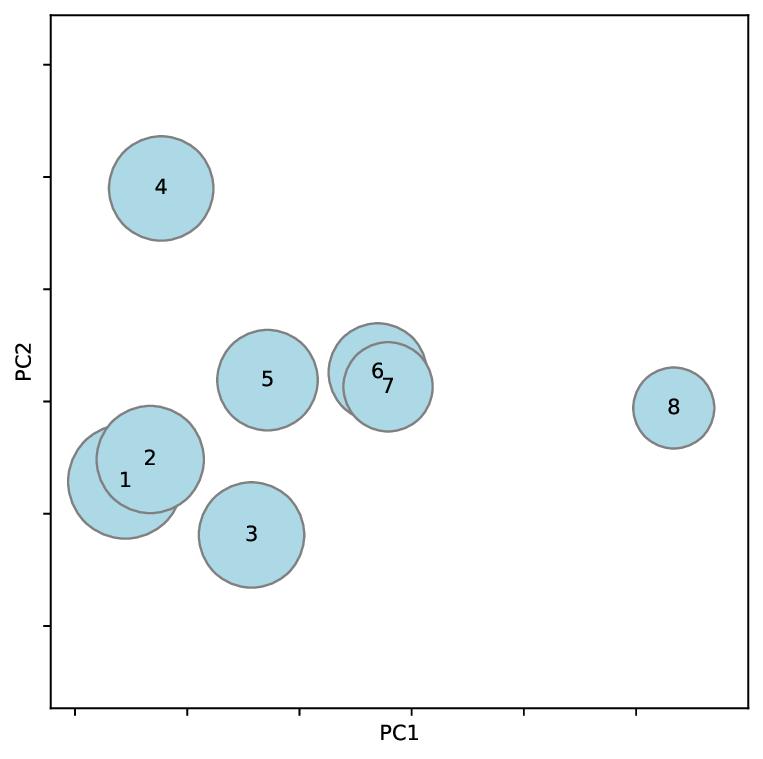}
  \end{minipage}
\caption{LDA, NMF, and BERTopic topic clusters}
    \label{fig:intertopic}
\end{figure*}

\textbf{Topic models}, which model the distribution of how topics appear within documents, have long been proposed as tools for computational social science and history \citep{blei2003latent, blei2006correlated}. In particular, \textit{dynamic topic models} \citep{blei2006dynamic} can be used to represent a distribution of topics that changes over time. However, experts have been critical of the practicality of topic models for research, noting the difficulty in quantitative evaluation \citep{chang_reading_2009}, handling documents with figurative language \citep{rhody2012topic}, and the difficulty in distinguishing novel, meaningful insights from noise \citep{goldstone_2014_quiet}.

Traditionally, topic models have used a Latent Dirichlet Allocation (LDA), a Bayesian statistical model that explains observed text using unseen groups of documents \citep{blei2003latent}. Recent research has proposed an alternate strategy for topic modeling, where neural embeddings are used to create topic clusters \citep{grootendorst2022bertopic}. In this work, we compare the effectiveness of this neural approach with traditional statistical techniques, hypothesizing that this approach will be more robust to noise and reduce the effort needed to create useful topic models.

Due to the cultural and political dominance of the Ancient Romans, there exists an extensive surviving corpus of Roman literature, with roughly 220,000 surviving texts and inscriptions \citep{clackson2011classical}. Many of these texts are individually well-studied, but holistic analysis of the corpus remains difficult. We experiment with various techniques for dynamic topic modeling over this entire corpus, which is noisy and broad in scope, to determine which approach performs best at producing useful results, while requiring minimal hyperparameter tuning.

We make the following contributions:
\begin{enumerate}
    \item We produce the first (to our knowledge) dynamic topic model spanning all surviving Roman texts, and create visualizations demonstrating clear historical trends.
    \item We find that the neural embedding approach produces a more readily interpretable topic distribution compared to traditional LDA approaches.
    \item We demonstrate that quantitative metrics for evaluating topic models do not necessarily align with human judgments, and better strategies for evaluating topic models are needed.
\end{enumerate}

Our code is available on GitHub.\footnote{link omitted for anonymity}

\section{Related Work}
Dynamic topic modeling has been used across disciplines to explore the evolution of bodies of literature such as nineteenth-century British Parliament debates \citep{Guldi2019ParliamentsDA}, medical publications about sepsis \citep{DoranBostwick201821481Y}, and Twitter data during the COVID-19 pandemic \citep{Bogdanowicz2022DynamicTM}. However, such works (and the majority of similar ones) use the traditional LDA model for dynamic topic modeling, while we utilize a modern neural clustering technique. Additionally, to our knowledge, this is the first work to apply dynamic topic models across the entire corpus of a historical state.

There has been growing interest in NLP approaches for ancient languages including Latin, evidenced by workshops such as LT4HALA \citep{lt4hala-2022-language}. Previous work in Latin NLP has included language modeling \citep{bamman2020latin}; lemmatization, part-of-speech tagging, and feature identification \citep{sprugnoli-etal-2022-overview}; and automatic dating of Latin texts \citep{allen_dlt2}.

\section{Data}
We use the Latin Library corpus as our source for Latin texts \citeplanguageresource{latinlibrary}. We date documents according to the information gathered in \citep{allen_dlt2}, although many of these dates are estimated years based on historical knowledge. After removing documents that did not have known dates, the corpus consisted of 1,350 plain-text documents, including histories, poems, plays, inscriptions, and speeches. Documents spanned from roughly 449 BC (\textit{The Twelve Tables}) through AD 600 (\textit{The Epistles of Gregory the Great}), which roughly covers the time span from the founding of Rome to its fall.\footnote{We chose not to include the majority of post-Roman literature, which includes many Latin texts but is often considered as a distinct corpus by classical scholars.} For authors with many small texts (poems and inscriptions), multiple texts were combined in a single document, whereas very long texts (such as the \textit{Aeneid}) were split into several documents.

As Latin is a highly inflectional language, preprocessing is essential to extract meaningful topic words. We tokenized, lemmatized, and removed stop words using the Classical Language Toolkit (CLTK) \citep{johnson-etal-2021-classical}.

\section{Methodology}
We trained and compared three different models for dynamic topic modeling: a Latent Dirichlet Allocation (LDA) model, as in \citet{blei2006dynamic}; a model based on Non-negative Matrix Factorization (NMF), and a neural, transformer-based model using BERTopic \citep{grootendorst2022bertopic}.\footnote{Implementation details and hyperparameters will be provided in an appendix in the final version.}


\subsection{LDA Model}
The LDA model follows the approach used in \citet{blei2006dynamic} closely, and uses Gensim\footnote{\url{https://radimrehurek.com/gensim/}} \citep{rehurek_lrec} for implementation.

The LDA dynamic topic model extends a static topic model, modeling each time slice as a static topic model. The parameters for the distribution at each time step are modeled with a state space distribution, capturing how the distribution of topics evolves over time. We separated documents into ten timesteps, finding through qualitative observation that this was the most effective setting across models.

\subsection{NMF Model}
Non-negative Matrix Factorization (NMF) is a linear algebra technique often used for extracting latent information. We reimplement the approach used in \citet{greene_exploring_2017} for dynamic topic modeling. In this approach, we run NMF on each timestep, factoring the matrix of documents and words into a matrix of documents to topics and topics to terms. Then, we run NMF again, using the terms for the topics at each timestep as the "documents". 

\subsection{BERTopic Model}
BERTopic \citep{grootendorst2022bertopic} uses neural embeddings of documents to create clusters of related documents, from which topics are extracted. For dynamic topic modeling, BERTopic uses the static topic model as a global distribution. We compute a topic distribution using the same topics for each time step and fine-tune this representation by averaging each time step globally and evolutionarily (with the previous time step).

\subsection{Evaluation}
Following the approach used in \citet{ocallaghan_analysis_2015}, we use several quantitative metrics to evaluate topic models. The \textsc{TC-Embed} metric scores \textit{topic coherence} by taking the average cosine similarity between embeddings for the topic ten words in a topic. The \textsc{Mean Pairwise Jaccard similarity} measure scores topic generality by taking the average ratio of shared terms to total terms for a pair of topics. A good topic model is thought to have high coherence and low generality.

\section{Results}
\vspace{-0.5 cm}
\begin{table}[!h]
  \def\arraystretch{1.5}

    \centering
    \begin{tabular}{| c|c c|}
    \hline
        Model & TC-Embed & MPJ \\
        \hline
        LDA & 0.209 & \textbf{0.016} \\
        NMF & \textbf{0.307} & 0.369 \\
        BERTopic & 0.189 & 0.065 \\
        \hline
    \end{tabular}
    \caption{Topic Coherence (TC-Embed) and Generality (MPJ) Measures, averaged over top-5 topics}
    \label{tab:topic_coherence}
\end{table}

The coherence and generality scores are given in \autoref{tab:topic_coherence}, showing that the LDA model minimalizes generality best, while the NMF model has the highest topic coherence, while the BERTopic model underperforms in both. However, prior research has suggested that these quantitative metrics often do not line up with human intuition about the quality of the model, particularly for dynamic topic models. Thus, we also evaluate the usefulness of the topic distributions from each model. \footnote{The full list of terms for each topic will be provided in an appendix in the final version}


\subsection{Intertopic Distance Map}
First, we analyzed the distribution of topics regardless of time. We use LDAvis and principal component analysis (PCA) to create a graph showing the distribution of topics for the two static topic models \citep{Sievert2014LDAvisAM}, presented in \autoref{fig:intertopic}. 

\subsection{Topics over time}
\begin{figure*}[!t]
    \centering
    \includegraphics[width=\textwidth]{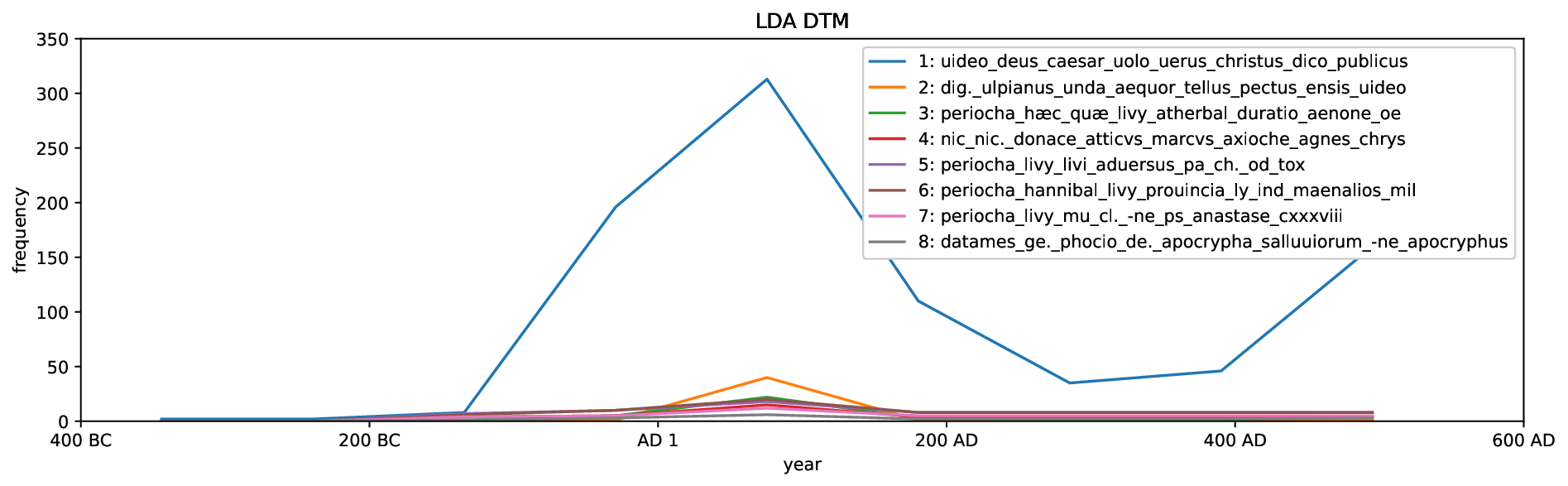}
    \includegraphics[width=\textwidth]{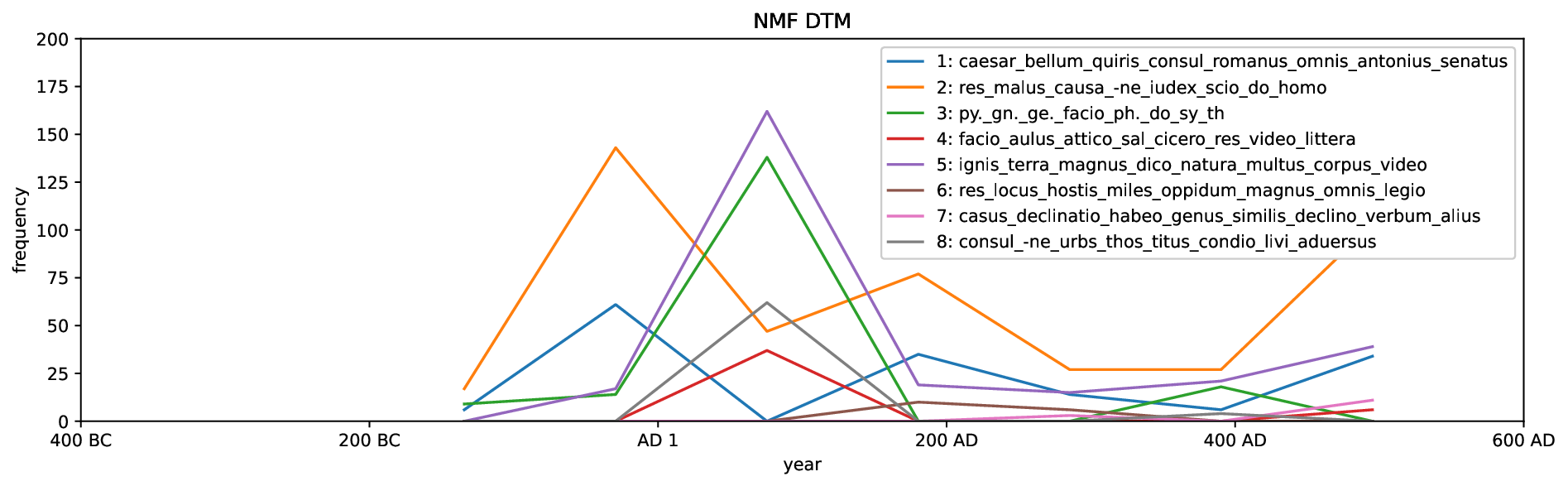}
    \includegraphics[width=\textwidth]{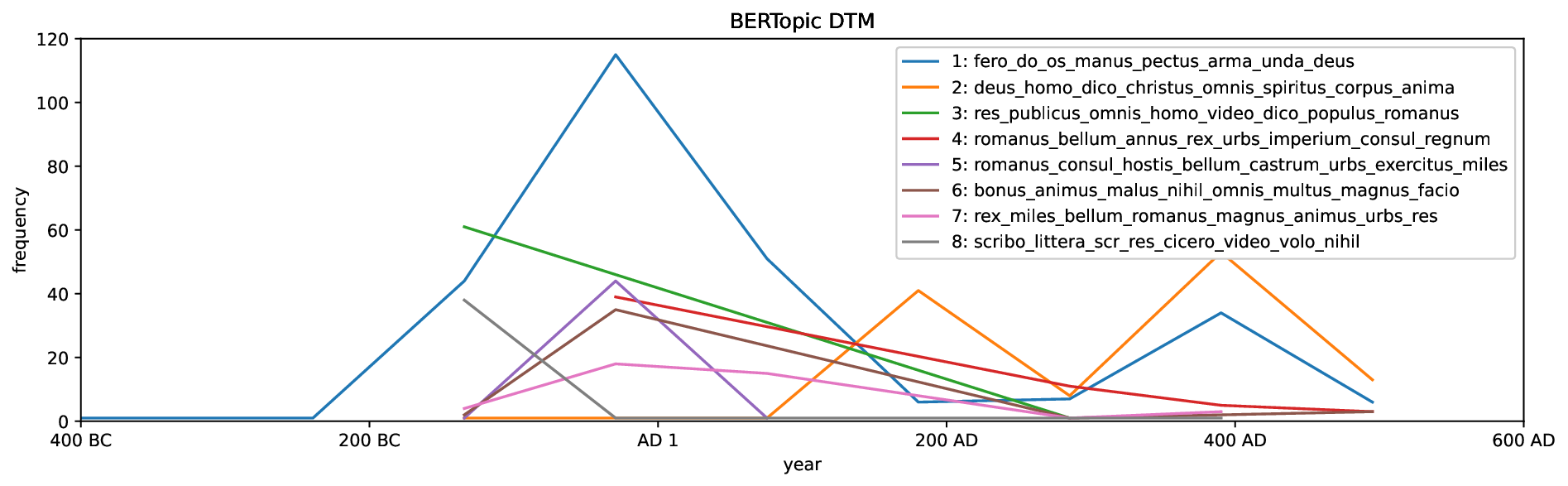}
    \caption{LDA, NMF, and BERTopic topic frequency (\# documents) over time}
    \label{fig:overtime}
\end{figure*}

We calculate the frequency of topics at each time step by calculating the number of documents with each topic as its most likely topic and  visualize the popularity of topics over time. These results are presented in \autoref{fig:overtime}.

\section{Discussion}
Though the LDA and NMF models outperform based on quantitative metrics, we can see in \autoref{fig:intertopic} and \autoref{fig:overtime} that their insights were limited, particularly for the LDA model.

The LDA intertopic map shows that Topic 1 and its cluster of topics were very general and dominant, and as a result \autoref{fig:overtime} does not reveal much of interest about the topics' frequency over time. The LDA model had difficulty distinguishing meaningful topic words from noise. We experimented with modifying hyperparameters (prior probabilities, chain variance) but were unable to make improvements.

The NMF model produced a wider range of topics, spread out and changing over time. However, most of the topic terms are difficult to interpret as meaningful, and the generality score indicates that the topics are mainly composed of common Latin terms. One promising topic is Topic 5, with terms such as \textit{ignis} ("fire") and \textit{magnus} ("great"), which peaks around AD 100 and could be related to the Great Fire of Rome in AD 64.

The results for the BERTopic model are insightful and useful, despite the quantitative measures, and the model requires minimal configuration. The intertopic map (\autoref{fig:intertopic}) shows a spread of evenly distributed topics, and the topics over time graph (\autoref{fig:overtime}) demonstrates clear trends in topic popularity, described below. BERTopic benefits from utilizing embedded document representations to create meaningful clusters of any shape, while the LDA model considers all documents equally. 

\subsection{Topic 2: Christianity in Rome}
One particularly clear and insightful topic is topic 1, which includes terms such as \textit{deus} ("god"), \textit{homo} ("man"), \textit{christus} ("Christ"), and \textit{spiritus} ("breath, spirit"), and is clearly associated with Christian religious writings. The topic peaks twice: once around A.D. 200, after Christianity was first established and introduced in Rome, and again around A.D. 400, after the Edict of Milan and the establishment of Christianity as the official religion of the Roman Empire.

\subsection{Topic 3: The fall of the Republic}
Another interesting topic is topic 3, including terms such as \textit{res publicus} ("republic"), \textit{populus} ("the people"), and \textit{romanus} ("Roman"). This topic seems to be associated with writings and speeches about Republican values such as Cicero's orations. We can see that the topic is most popular around 100 B.C. and decreases in popularity through A.D. 300, likely a reflection of the fall of the Roman Republic (44 B.C.) and its associated republican ideals.

\subsection{Comparison}
The results from the LDA and NMF models were consistent with many of the criticisms from humanities and social science experts: the models did not extract particularly useful information, and the quantitative metrics did not align with human judgments. We believe that with extensive hyperparameter search, we could achieve comparable results to the BERTopic model, but this effort may not be worth the resulting value to a historian.

In contrast, the BERTopic model mitigates many of these issues, with a more robust method for clustering documents that produces insights with minimal effort. However, it does require an existing model for creating document embeddings, which may not always be available for many low-resource languages. Future work could explore using a state space to model distributional parameters over time while incorporating neural embeddings to guide topic extraction.

\section{Conclusion}
In this work, we trained three different dynamic topic models over documents spanning the lifetime of the Roman Republic and Roman Empire. We find that the traditional LDA and NMF models suffer from common issues with topic modeling: they may be difficult to interpret in a useful way, they are sensitive to noisy data, and they often require parameter search to find optimal configurations. In contrast, the BERTopic model is able to achieve insightful results with minimal effort, producing trends that line up with historical intuitions. We believe this topic model could be useful in certain scenarios; for example, a Roman historian might be interested in how a particular author's distribution of topics relates to the overall historical trends. 

While this approach shows clear potential in the use of topic models for historical analysis, it does not solve all of the issues. Interpretation is still critical and difficult, and there is something of a paradox: it is difficult to tell if a trend in a topic model is meaningful and accurate, without already knowing that the trend exists. Nevertheless, we hope that newer approaches can make topic modeling more robust and performant as one of many tools for discovering insights and performing research.




\section{Ethics}
We hope that through more sophisticated methods for topic modeling, we can aid historians and social scientists in their research. However, topic modeling and other computational approaches for humanities should not supplant human experts, who are able to investigate research questions with more care and nuance than an automated method. Topic models, and other forms of statistical text summarization, tend to extract majority ideas and topics. Topic models should be used to gain holistic information about trends and topics, but should not be used to overlook minority opinions.

This work uses Latin text which is now in the public domain, but any work utilizing data for living languages should ensure they have proper permission to use the data, especially with indigenous and endangered languages. Communities should always have control over the data they produce, and language should be seen as a cultural and personal artifact, not merely data for conducting research.

Finally, models that require high computational costs, particularly neural models, use large amounts of energy and carry an environmental cost \citep{strubell_energy_2020}. 
\section{Bibliographical References}\label{sec:reference}

\bibliographystyle{lrec-coling2024-natbib}
\bibliography{custom}

\section{Language Resource References}
\label{lr:ref}
\bibliographystylelanguageresource{lrec-coling2024-natbib}
\bibliographylanguageresource{languageresource}

\end{document}